\documentclass[10pt,twocolumn,letterpaper]{article}

\usepackage[pagenumbers]{cvpr}%

\usepackage{iftex}
\usepackage{array}%
\ifxetex
  \usepackage{fontspec}
\fi
\usepackage{caption}
\usepackage{booktabs}
\usepackage{multirow}
\usepackage{graphicx}
\usepackage{amsmath,amssymb}
\usepackage{lipsum}%

\newcommand{\method}{MonoCal}

\definecolor{calblue}{HTML}{0072B2}
\definecolor{oursorange}{HTML}{D55E00}
\definecolor{relgray}{HTML}{6E6E6E}
\newcommand{\lbl}[1]{#1}%

\newcommand{\sig}{\sigma}
\newcommand{\rmed}{r_{\mathrm{med}}}
\newcommand{\rrms}{r_{\mathrm{rms}}}
\newcommand{\ence}{\mathrm{ENCE}}
\newcommand{\encemed}{\mathrm{ENCE}_{\mathrm{med}}}
\newcommand{\sience}{\mathrm{SI\text{-}ENCE}}
\newcommand{\siencemed}{\mathrm{SI\text{-}ENCE}_{\mathrm{med}}}

\definecolor{cvprblue}{rgb}{0.21,0.49,0.74}
\usepackage[pagebackref,breaklinks,colorlinks,allcolors=cvprblue]{hyperref}

\title{Calibrating the Uncertainty of Self-Supervised Monocular Depth without Labels}

\author{Jun Seok Kim \qquad Rui Huang$^{*}$\\
The Chinese University of Hong Kong, Shenzhen\\
{\tt\small junseokkim@link.cuhk.edu.cn, ruihuang@cuhk.edu.cn}}

\newcommand{\teaserpanela}{MonoProb~\cite{marsal2024_monoprob}. The scale is the error divided by the uncertainty over all
  pixels, and the shape is whether the curve follows the line.}
\newcommand{\teaserpanelb}{The scale before and after scale calibration on four other networks on KITTI, one of them from
  Poggi~\etal~\cite{poggi2020_uncertainty}.}

\newcommand{\methodbars}{}

\newcommand{\teaserpanelc}{On two other datasets.}

\begin{document}
\twocolumn[{%
\renewcommand\twocolumn[1][]{#1}%
\vspace*{-20pt}%
\maketitle
\vspace{-20pt}%
\begin{center}
  \centering
  \captionsetup{type=figure}
  \includegraphics[width=\textwidth]{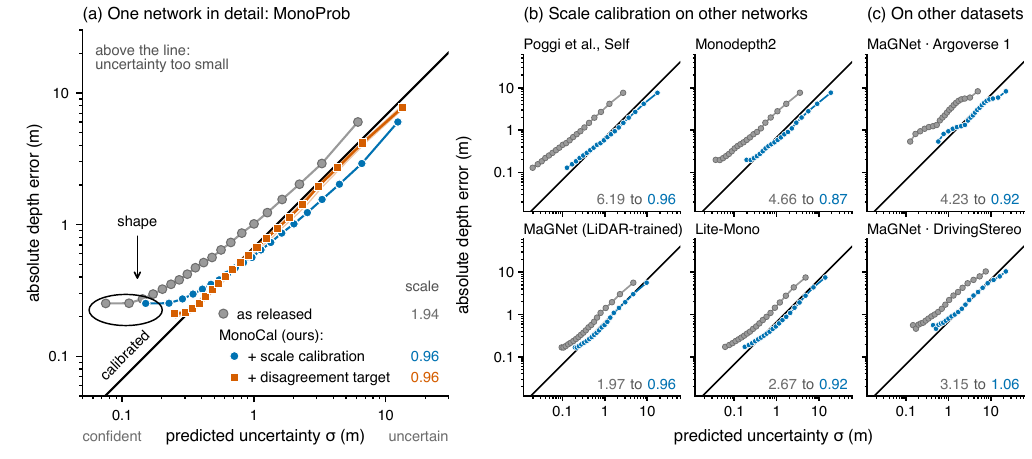}
  \captionof{figure}{\textbf{The uncertainty of self-supervised depth is miscalibrated in scale and shape, and \method{} corrects both without
  labels.} \textbf{(a)}~\teaserpanela{} \textbf{(b)}~\teaserpanelb{} \textbf{(c)}~\teaserpanelc{}}
  \label{fig:teaser}
\end{center}%

}]
{\renewcommand{\thefootnote}{*}\footnotetext{Corresponding author.}}%
\begin{abstract}
Self-supervised monocular depth networks can predict a per-pixel uncertainty, which is mainly evaluated by how well it ranks
pixels according to their error. Such metrics are insensitive to the magnitude of the uncertainty, so an uncertainty half as large
as the actual error obtains the same score. In this paper, we investigate whether the predicted uncertainty is as large as the
depth error, both on average (\emph{scale}) and across confident and uncertain, near and far pixels (\emph{shape}). On KITTI, the
scale is miscalibrated for every uncertainty estimate we evaluate, and the error of MonoProb is nearly twice its predicted uncertainty. We
propose \method{}, which corrects both the scale and the shape without ground-truth depth. \emph{Scale calibration} multiplies the
uncertainty by a single constant measured from stereo pairs, requiring no retraining, and this constant closely matches the one
fitted with ground truth. Since a constant cannot change the shape, we further retrain the uncertainty to predict where the
depth of the network disagrees with that of a second self-supervised network. Experiments on KITTI show that \method{} achieves better calibration
across uncertainty levels and depth ranges than rescaling with ground truth, while still using a single image at inference. Code will be made publicly available.
\end{abstract}

\begin{table}[!h]
  \centering\footnotesize
  \caption{\textbf{How we measure the magnitude of an uncertainty estimate.} The ratio is the absolute depth error divided by the
  predicted uncertainty. It is 1 when the uncertainty is calibrated.}
  \label{tab:measures}
  \setlength{\tabcolsep}{2.2pt}
  \scriptsize
  \begin{tabular}{l l >{\raggedright\arraybackslash}p{0.66in} >{\raggedright\arraybackslash}p{0.47in} >{\raggedright\arraybackslash}p{0.52in} >{\raggedright\arraybackslash}p{0.40in}}
    \toprule
    & & is the uncertainty right & ratio taken over & reported as & shown in \\
    \midrule
    \multicolumn{2}{l}{scale} & on average? & all pixels & $\rmed$, $\rrms$ ($\to 1$) & Fig.~\ref{fig:teaser}, numbers \\
    \midrule
    \multirow{5}{*}{shape $\left\{\rule{0pt}{5.6ex}\right.$} & consistency & for confident and uncertain pixels alike? & bins by uncertainty & $\encemed$, $\ence$ ($\downarrow$) & Fig.~\ref{fig:teaser}, curves \\
    \addlinespace[5pt]%
    & adaptivity  & for near and far pixels alike? & bins by depth & $\siencemed$, $\sience$ ($\downarrow$) & Fig.~\ref{fig:profile} \\
    \bottomrule
  \end{tabular}
\end{table}

\section{Introduction}
\label{sec:intro}

Systems that fuse depth measurements rely on the magnitude of the predicted uncertainty. Dense mapping systems weight or reject
each depth measurement according to an uncertainty that is specified by hand as a function of distance or of the
sensor~\cite{oleynikova2017_voxblox,nguyen2012_kinectnoise,millane2023_nvblox,boche2025_okvis2x}, learned by a supervised
network~\cite{laidlow2019_deepfusion,boche2025_okvis2x}, or derived from bundle adjustment~\cite{rosinol2023_probvolfusion}. A learned
uncertainty can replace these models only if it has the correct magnitude. However, self-supervised monocular depth networks that predict
an uncertainty $\sig$, interpreted as the standard deviation of the predicted depth, are typically evaluated using sparsification
metrics such as the area under the sparsification error (AUSE)~\cite{ilg2018_uncertainty,poggi2020_uncertainty}. These metrics assess
whether $\sig$ ranks the pixels according to their error. Since they are invariant to any rescaling of $\sig$, a $\sig$ that is half
or twice the size it should be obtains the same score~\cite{gustafsson2020_evaluating,marsal2024_monoprob}.

In this paper, we measure this magnitude using a single ratio, the absolute depth error divided by the predicted uncertainty, which
equals 1 for a calibrated $\sig$ and 2 for a $\sig$ that is half the size it should be. We examine two properties of this ratio
(Tab.~\ref{tab:measures}). The \emph{scale} describes whether the ratio is 1 over all pixels, and the \emph{shape} describes whether
the ratio is the same for pixels of small and large $\sig$ and for near and far pixels~\cite{pernot2023_adaptivity}. For
MonoProb~\cite{marsal2024_monoprob}, which predicts the uncertainty as a standard deviation in meters, the error is nearly twice the
predicted uncertainty over all pixels (Fig.~\ref{fig:teaser}a) and more than three times in the nearest depth range, indicating that the
network is overconfident. Moreover, the scale is miscalibrated for every uncertainty estimate we evaluate on KITTI, including
published self-supervised networks, three other backbones and systems trained with depth labels, by factors between 0.6 and 6
(Fig.~\ref{fig:levelall}). For self-supervised networks, the training objective does not constrain this magnitude, since the network is
trained to reconstruct images rather than to regress depth~\cite{poggi2020_uncertainty} and no loss term compares $\sig$ with a
depth error.

Existing systems correct the magnitude using a constant fitted against a reference depth, on a network trained with labels
(Sec.~\ref{sec:related}). However, this constant is not universal. When measured by the same rule, it varies across networks and
datasets (Fig.~\ref{fig:levelall}), which has also been observed in supervised regression~\cite{laves2020_wellcalibrated}. The
constant therefore has to be measured where $\sig$ is used, and for self-supervised depth this is a setting without labels.

To address this, we propose \method{}, which corrects the scale and the shape without labels using two components. The first
component, \emph{scale calibration}, measures the constant. We obtain a depth from a stereo pair by semi-global
matching~\cite{hirschmuller2008_sgm} and compare it with the depth of the network. The median ratio of this residual to $\sig$ then
determines a single constant. This procedure requires neither ground-truth depth nor retraining of the network. The stereo pair is
required only to measure the constant, while inference still uses a single image. On KITTI, the constant is within $\times 1.08$ of
the ground-truth constant for 16 of 17 uncertainty estimates (Fig.~\ref{fig:levelall}, Fig.~\ref{fig:teaser}a).

\begin{figure*}[t]
  \centering
  \includegraphics[width=\textwidth]{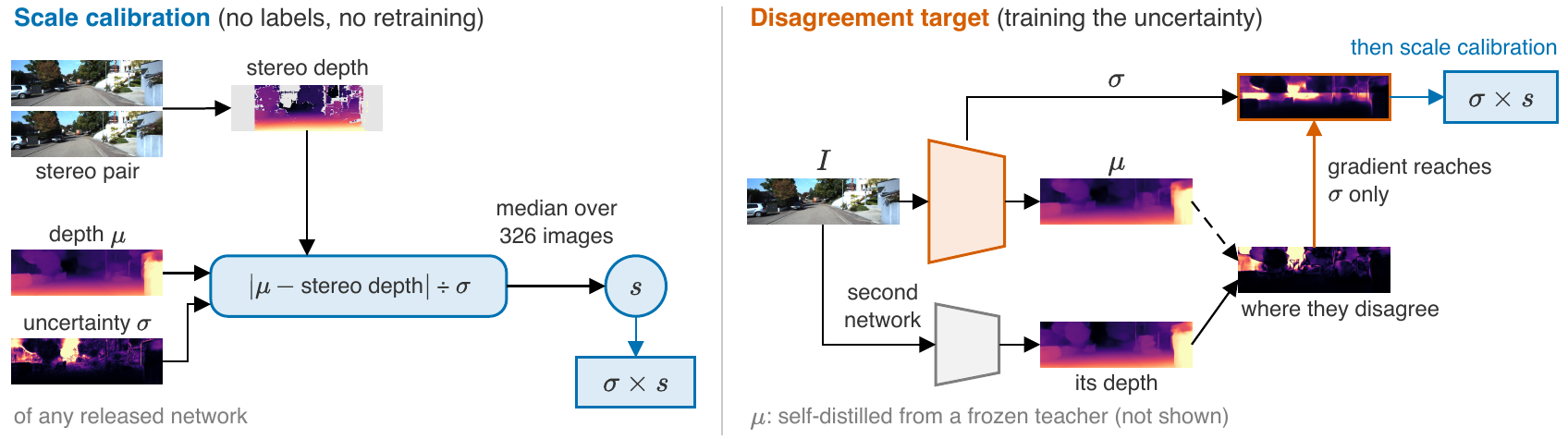}
  \caption{\textbf{Overview of \method{}.} \emph{Scale calibration} (blue) measures one constant $s$ from the
  residual between the network's depth $\mu$ and a stereo depth, and multiplies the uncertainty $\sig$ by it. The
  \emph{disagreement target} (orange) trains $\sig$ on where $\mu$ disagrees with the depth of a second self-supervised network.
  \method{} applies the target, then the calibration. Inference uses one image.\methodbars{}}
  \label{fig:method}
\end{figure*}

However, a constant can only correct the scale. On the reliability diagram in Fig.~\ref{fig:teaser}(a), a constant shifts the
curve horizontally without changing its slope. Similarly, it shifts the entire profile along depth, so that MonoProb still
underestimates the uncertainty in the nearest depth range even with the ground-truth constant (Fig.~\ref{fig:profile}). The second
component, the \emph{disagreement target}, is designed to correct this shape. We retrain $\sig$ to predict where the depth of the
network disagrees with that of a second, independently trained self-supervised network of a different architecture, and then apply
scale calibration (Fig.~\ref{fig:method}). On KITTI, the ratio in the nearest depth range becomes 0.96, and the calibration errors
across uncertainty levels and across depth are lower than those of the ground-truth constant (Tab.~\ref{tab:main}). The consistency
gain generalizes to two of the three other backbones (Tab.~\ref{tab:backbones}).

\noindent\textbf{Contributions.}
\begin{enumerate}\setlength{\itemsep}{0pt}
  \item We measure the magnitude of self-supervised depth uncertainty by separating scale from shape, and show that the scale is
        miscalibrated for every uncertainty estimate we evaluate, by a factor that varies across networks and datasets.
  \item We propose \emph{scale calibration}, a single label-free constant measured from stereo residuals without retraining. On
        KITTI, this constant agrees with the ground-truth constant within $\times 1.08$ for 16 of 17 uncertainty estimates.
  \item We propose the \emph{disagreement target}, which trains $\sig$ on the disagreement with a second self-supervised network.
        On KITTI, it reduces the calibration errors across uncertainty levels and across depth below those of the ground-truth
        constant.
\end{enumerate}

\section{Related Work}
\label{sec:related}

\noindent\textbf{Uncertainty in self-supervised depth.}
Self-supervised monocular depth is trained by view synthesis, from stereo pairs~\cite{godard2017_monodepth} or
video~\cite{godard2019_monodepth2}. Per-pixel uncertainty was introduced in this setting as the width of a Laplacian over the photometric
residual~\cite{klodt2018_supervising,yang2020_d3vo}, by applying the heteroscedastic loss of Kendall and
Gal~\cite{kendall2017_uncertainties} to the reprojection error. In this formulation, the magnitude of the uncertainty is determined by the
photometric error rather than by a depth error, and training a variance with a Gaussian likelihood has known failure
modes~\cite{seitzer2022_pitfalls,stirn2023_faithful}.
Poggi~\etal~\cite{poggi2020_uncertainty} compared empirical, predictive and Bayesian estimators using the ranking metrics AUSE and
AURG, and found self-teaching, in which the depth of a second network serves as the target, to be among the most useful.
MonoProb~\cite{marsal2024_monoprob} samples the photometric loss over a predicted distribution and offers a self-distilled variant
in which a student learns the depth distribution of a teacher. These works evaluate $\sig$ through
sparsification~\cite{ilg2018_uncertainty}, and ConFiDeNet~\cite{susladkar2026_confidenet}, a recent supervised system that compares
against MonoProb, additionally reports the ARU and RMSU metrics of MonoProb.
However, sparsification curves are invariant to a monotone rescaling of $\sig$, and ARU and RMSU, which are pointwise distances
between $\sig$ and the error, are minimized by a $\sig$ below calibration (Sec.~\ref{sec:exp_main}).
A second depth has previously been used as a target, namely SfM depth in Klodt and Vedaldi~\cite{klodt2018_supervising} and a
teacher of the same architecture in self-teaching and in the self-distillation of MonoProb. In all three cases, this target also
supervises the depth. In contrast, we use the disagreement between two networks of different architectures as the target of $\sig$
alone.

\noindent\textbf{Calibration of regression uncertainty.}
Pernot~\cite{pernot2023_adaptivity} defines three calibration targets for a predicted $\sig$, namely average calibration (our
\emph{scale}), consistency (calibration as a function of $\sig$ itself) and adaptivity (calibration as a function of the input), and
shows that consistency does not imply adaptivity. We refer to the last two as the \emph{shape}. The expected normalized calibration error of
Levi~\etal~\cite{levi2022_evaluating} measures consistency as the mean over $\sig$ bins of $|\mathrm{RMSE}/\mathrm{RMV} - 1|$.
Rescaling $\sig$ by a constant fitted on labeled data is a standard recalibration step. For example, $\sig$
scaling~\cite{levi2022_evaluating,laves2020_wellcalibrated} fits a single scalar by the likelihood on a validation set, similar to
temperature scaling for classification~\cite{guo2017_calibration}, and isotonic recalibration~\cite{kuleshov2018_calibrated} fits a
monotone map. Calibration without target labels has been studied for classification under domain shift~\cite{hu2023_pseudocal}. Our
work adds a way to measure the constant without labels and a training target that corrects what the constant cannot.
In depth estimation, f-Cal~\cite{bhatt2021_fcal} reports the expected calibration error of a label-trained network.
McCraith~\etal~\cite{mccraith2020_calibrating} calibrate the metric scale of self-supervised depth rather than the magnitude of its uncertainty.
Gustafsson~\etal~\cite{gustafsson2020_evaluating} note that a perfect AUSE is compatible with a consistently under-estimated $\sig$.
Hornauer~\etal~\cite{hornauer2025_revisiting} compare error and uncertainty per bin on self-supervised Monodepth2 and MonoViT, but
first apply min--max normalization to both, which again removes the magnitude of $\sig$.
Similarly, Landgraf~\etal~\cite{landgraf2025_critical} combine five uncertainty estimators with Depth Anything~V2~\cite{yang2024_depthanythingv2}
and evaluate how well the uncertainty ranks the errors, not its magnitude.
To our knowledge, consistency and adaptivity have not been
measured for the $\sig$ of self-supervised depth.
Concurrently, Deng~\etal~\cite{deng2026_calibaudit} audit the confidence of seven feed-forward 3D reconstruction models using the
ratio of error to predicted uncertainty per confidence bin, and repair it post hoc using a two-constant power law fitted with
ground truth per backbone and dataset. In contrast, we evaluate self-supervised monocular $\sig$, measure the constant without labels and
also examine calibration along depth.

\noindent\textbf{How systems calibrate uncertainty in practice.}
Jung~\etal~\cite{jung2024_uncertaintyslam} fit a constant so that the mean Mahalanobis distance on a validation set is one.
MP-SfM~\cite{pataki2025_mpsfm} fits a constant to an SfM point cloud of another dataset and adds a depth-proportional floor.
Mono-SF~\cite{brickwedde2019_monosf} trains a recalibration network on held-out ground truth, and
Tao~\etal~\cite{tao2022_lidarcompletion} train $\sig$ with ground truth and choose the rejection threshold by mapping performance.
All four build on networks trained with labels, and three of them recalibrate $\sig$ before use.
OKVIS2-X~\cite{boche2025_okvis2x} states that heuristic uncertainty models are not sufficient for learned depth.

\section{Measuring the Magnitude of Uncertainty}
\label{sec:measures}

For a pixel with predicted depth $\mu$, predicted uncertainty $\sig$ and ground-truth depth $d_{\mathrm{gt}}$, we define the error as the
absolute depth error $e = |\mu - d_{\mathrm{gt}}|$. Since a calibrated $\sig$ is as large as the errors it accompanies, we compute
the ratio $r$ of error to predicted $\sig$ over a set of pixels using two statistics. The first,
$\rmed = \mathrm{median}(e/\sig)\,/\,0.6745$, takes the per-pixel ratio first (the median absolute deviation of a Gaussian is
$0.6745\,\sig$). The second, $\rrms = \mathrm{RMSE}/\mathrm{RMV}$, divides the root-mean-square error by the root-mean-variance. For
both statistics, $r = 1$ corresponds to a calibrated $\sig$ and $r = 2$ to a $\sig$ that is half the size it should be. The median
statistic remains exact for a calibrated Gaussian $\sig$ that varies from pixel to pixel within the set. In contrast, the RMS
statistic corresponds to what a downstream system assuming Gaussian errors observes, but it is sensitive to the tails and can
therefore be unreliable under heavy-tailed distributions~\cite{pernot2024_heavytails}. We therefore report both statistics and
compare results within the same statistic.

The \emph{scale} of $\sig$, not to be confused with the scale of the depth, is defined as $r$ over all pixels, which corresponds to the average calibration of
Pernot~\cite{pernot2023_adaptivity}. The \emph{shape} describes how $r$ varies across pixels along two axes (Tab.~\ref{tab:measures}), which we call consistency and adaptivity following Pernot~\cite{pernot2023_adaptivity}. For \emph{consistency}, pixels are sorted by predicted $\sig$ into 20 bins of equal count, and $r$ is
computed for each bin. We summarize it by $\ence = \frac{1}{20}\sum_g |r_g - 1|$ using the RMS
statistic~\cite{levi2022_evaluating} and by $\encemed$, the same mean using $\rmed$ per bin. Since each bin is compared with 1, a
miscalibrated scale also increases this error, and we therefore compare it after applying the constant. For \emph{adaptivity}, the
20 bins are instead formed by ground-truth depth, since the same $\ence$ measures adaptivity when binned by a variable other than
$\sig$~\cite{pernot2023_adaptivity}. We summarize adaptivity by
\begin{equation}
  \sience = \min_c \frac{1}{20}\sum_k |c\,r_k - 1| ,
  \label{eq:sh}
\end{equation}
which is the depth-binned $\ence$ after applying the best single constant, an oracle $\sig$
scaling~\cite{laves2020_wellcalibrated} chosen on the evaluation images in the same way for every method. We call it the scale-invariant $\ence$ ($\sience$), where the scale is that of $\sig$, and write $\siencemed$ when $r_k$ is $\rmed$. It is therefore invariant
to any rescaling of $\sig$ and measures only what a constant cannot remove.

\noindent\textbf{Observations.}
On KITTI, the scale is miscalibrated for every uncertainty estimate we evaluate, by factors between 0.6 and 6
(Fig.~\ref{fig:levelall}). For MonoProb, the error is 1.94 times the predicted uncertainty, and the shape is also miscalibrated, as
the ratio is larger for small $\sig$ than for large $\sig$ (Fig.~\ref{fig:teaser}a) and largest in the nearest depth range
(Fig.~\ref{fig:profile}).

\section{Method}
\label{sec:method}

\method{} consists of two components (Fig.~\ref{fig:method}): \emph{scale calibration}, a constant that can be applied to any
pretrained network (Sec.~\ref{sec:scale}), and the \emph{disagreement target}, a training target for what the constant cannot change
(Sec.~\ref{sec:cannot} and~\ref{sec:target}).

\subsection{Scale Calibration}
\label{sec:scale}

With ground truth, the scale could be corrected by fitting the constant that yields $\rmed = 1$ on labeled images. We measure the
same constant using a stereo pair instead of the label. For each validation image, we obtain a depth $d_S$ from the stereo pair by
semi-global matching~\cite{hirschmuller2008_sgm}, which has been used as a proxy label for monocular
depth~\cite{tosi2019_monoresmatch}. We align $d_S$ to the depth $\mu$ of the network by the per-image median ratio and measure the
constant at the pixels where $\sig$ is evaluated, i.e., the LiDAR-covered pixels of the standard evaluation crop. Only the positions
of these pixels are used, not their depth values. Without the positions, the constant of \method{} changes from $\times 2.26$ to
$\times 2.22$, i.e., by less than 2\,\%. Other pixel sets are compared in Tab.~S5 of the supplementary material.
The stereo residual $e_S = |\mu - d_S|$ then serves as a substitute for the error, and the constant is computed as
\begin{equation}
  s = \exp\!\Big(\operatorname*{median}_{p}\,\log\frac{e_{S,p}}{\sig_p}\Big) \Big/\, 0.6745
  \label{eq:s}
\end{equation}
over all evaluated pixels $p$ of the 326 validation images, which is the value that makes the median of $e_S/(s\,\sig)$ equal to that
of a Gaussian. The calibrated uncertainty is $s\,\sig$, while $\mu$ and the network remain unchanged.

After multiplying by $s$, the scale of $\sig$ with respect to the stereo residual is one by construction. The scale with respect
to the true error is also one exactly when the median of $\log(e_S/\sig)$ equals that of $\log(e/\sig)$, which requires the stereo
residual to be as large as the error in the median, but not for each pixel. Since stereo matchers introduce errors of their own,
and estimating their per-pixel confidence is a research field in itself~\cite{poggi2021_confidence}, whether this median condition
holds is an empirical question, which we examine in Sec.~\ref{sec:exp_general} for 17 uncertainty estimates. Stereo is required
only to measure $s$, while inference uses a single image.

\subsection{Limitation of a Single Constant}
\label{sec:cannot}

\begin{figure*}[t]
  \centering
  \includegraphics[width=\textwidth]{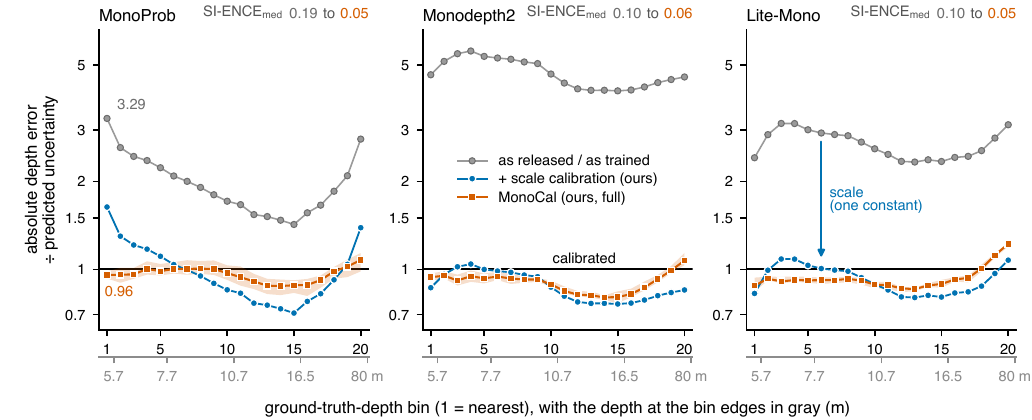}
  \caption{\textbf{A constant shifts the depth profile, and the disagreement target flattens it.} Absolute depth error divided by
  predicted uncertainty in 20 equal-count bins of ground-truth depth (1 is calibrated). Band: range over runs. On the two other
  backbones the change is smaller (Tab.~\ref{tab:backbones}).}
  \label{fig:profile}
\end{figure*}

Scale calibration multiplies every $\sig$ by the same number. On the reliability diagram of predicted $\sig$ against error
(Fig.~\ref{fig:teaser}a), which is drawn in log coordinates, this multiplication corresponds to a translation along $x$. The curve
is therefore shifted onto the calibration line, while its slope and its bends are preserved. A calibrated $\sig$ yields a slope of
1, since the error grows in proportion to $\sig$. In contrast, MonoProb has a slope of 0.74, both before and after applying any
constant. When the $\sig$ of MonoProb increases tenfold, the error increases only 5.5-fold, which means that $\sig$ overstates its
own differences and that the uncertainty of small-$\sig$ pixels is underestimated more than that of large-$\sig$ pixels.
The same holds along depth. A constant scales the entire depth profile of $r$ by a single factor, so correcting the middle ranges
leaves both ends underestimated, whereas correcting the nearest range makes the middle ranges overestimated
(Fig.~\ref{fig:profile}). By definition, $\sience$ also remains unchanged.
Consequently, the shape cannot be corrected by any constant, whether label-free or fitted with ground truth, and a different
$\sig$ is required.

\begin{table*}[t]
  \centering\footnotesize
  \caption{\textbf{Main results on KITTI with the MonoProb backbone.} \method{} (full) is the disagreement target followed by scale calibration. A constant changes neither the depth nor the
  ranking of pixels, so the first three rows share $\sience$, AUSE and abs\_rel. Bold: best label-free result. $\dagger$: lowest for a $\sig$ that is too small, so
  no value is bold.}
  \label{tab:main}
  \setlength{\tabcolsep}{3.9pt}%
  \begin{tabular}{l rr rr rr rr rr}
    \toprule
    & \multicolumn{2}{c}{scale ($\to 1$)} & \multicolumn{4}{c}{shape ($\downarrow$)} & \multicolumn{4}{c}{standard} \\
    \cmidrule(lr){2-3}\cmidrule(lr){4-7}\cmidrule(lr){8-11}
    & & & \multicolumn{2}{c}{consistency} & \multicolumn{2}{c}{adaptivity} & & & & \\
    \cmidrule(lr){4-5}\cmidrule(lr){6-7}
    uncertainty & $\rmed$ & $\rrms$ & $\encemed$ & $\ence$ & $\siencemed$ & $\sience$ & AUSE\,$\downarrow$ & abs\_rel\,$\downarrow$ & ARU$^{\dagger}$ & RMSU$^{\dagger}$ \\
    \midrule
    \textcolor{relgray}{MonoProb (released)} & 1.937 & 1.651 & 1.072 & 2.882 & 0.194 & 0.150 & 0.034 & 0.092 & 0.067 & 2.916 \\%
    \lbl{$+$ ground-truth constant (uses labels)} & \lbl{0.900} & \lbl{0.767} & \lbl{0.304} & \lbl{0.846} & \lbl{0.194} & \lbl{0.150} & \lbl{0.034} & \lbl{0.092} & \lbl{0.088} & \lbl{3.765} \\%
    \textcolor{calblue}{$+$ scale calibration (ours)} & \textbf{0.964} & \textbf{0.821} & 0.301 & 0.969 & 0.194 & 0.150 & 0.034 & 0.092 & 0.084 & 3.582 \\%
    \textcolor{oursorange}{\method{} (ours, full)} & 0.960 & 0.696 & \textbf{0.077} & \textbf{0.208} & \textbf{0.048} & \textbf{0.083} & \textbf{0.025} & \textbf{0.091} & 0.072 & 3.538 \\%
    \bottomrule
  \end{tabular}
\end{table*}

\subsection{Disagreement Target}
\label{sec:target}

To correct the shape, we retrain $\sig$ using a target that is available without labels, namely the disagreement between the depth
of the network and the depth of a second self-supervised network.%
Following MonoProb~\cite{marsal2024_monoprob}, a student network is retrained from ImageNet initialization by self-distillation
from the frozen MonoProb network, using the loss $L_{\mathrm{distill}}$, i.e., the negative log-likelihood of the residual with
respect to this teacher. We add a term of the same form, $L_\sig$, in which the teacher is replaced by
$d_{\mathrm{ext}}$, the depth of a second self-supervised network of a different architecture
(MonoViT~\cite{zhao2022_monovit}, public KITTI weights):
\begin{equation}
  L_\sig = \operatorname*{mean}_{p}\Big[\frac{(\mathrm{sg}(\mu_p) - d_{\mathrm{ext},p})^2}{2\,\sig_p^2} + \log \sig_p\Big].
  \label{eq:lsig}
\end{equation}
This term is computed after aligning $\mu$ and $\sig$ to $d_{\mathrm{ext}}$ by the per-image median ratio. The operator
$\mathrm{sg}$ stops the gradient through $\mu$, so that the second network affects only $\sig$ and the depth is trained by
$L_{\mathrm{distill}}$ alone. The total loss is $L_{\mathrm{distill}} + L_\sig$ (settings in Sec.~S1).
After training, $\sig$ is multiplied by the constant $s$ of Sec.~\ref{sec:scale}, while $\mu$ remains unchanged. In summary,
\method{} applies the disagreement target followed by scale calibration.

\noindent\textbf{Motivation.}
Disagreement between trained models has long been used to estimate generalization error~\cite{krogh1995_ensembleambiguity} and
predictive uncertainty~\cite{lakshminarayanan2017_ensembles}. Two networks of different architectures disagree where at least one of
them is inaccurate, which allows their disagreement to serve as a substitute for the error. However, the disagreement does not necessarily have the magnitude of the error. Two self-supervised networks agree with each other more than either agrees with LiDAR, and
before applying the constant, the retrained $\sig$ is further from calibration than that of MonoProb (Fig.~\ref{fig:levelall}).
The target therefore determines where $\sig$ is large, while scale calibration determines how large it is. Where both networks
make the same error, the target provides no signal, and the constant can only restore the magnitude.

\noindent\textbf{Cost.}
The student is retrained together with its depth, and the depth accuracy is unchanged or better (Tab.~\ref{tab:main}). A single
run takes about 3.7\,h on one RTX~2080~Ti, and the second network is required only during training.
We compare alternatives to this configuration in Tab.~\ref{tab:ablation}.

\section{Experiments}
\label{sec:experiments}

\subsection{Experimental Setup}
\label{sec:exp_setup}

\noindent\textbf{Data and protocol.}
We conduct our experiments on KITTI~\cite{geiger2013_kitti}. We use the Eigen split~\cite{eigen2014_depth} with improved
ground truth~\cite{uhrig2017_sparsity}, and split its 652 test images by drive into a validation set and a test set (326 images
each). The constants are estimated and the configuration is selected on the validation set, and all methods are evaluated on the
test set.
Similarly, on Argoverse~1 and DrivingStereo, we estimate the constants on validation logs and evaluate on test
logs, whereas nuScenes is evaluated on its validation set. All methods are evaluated on the same pixels, i.e., the official
evaluation mask with $10^{-3} < d_{\mathrm{gt}} < 80$\,m, in which the nearest LiDAR depth is about 3.7\,m. Following the standard
protocol for self-supervised monocular depth, $\mu$ is scaled per image by its median ratio to the ground truth, and $\sig$ is
scaled by the same factor.
The depth bins are fixed once and shared by all methods.

\noindent\textbf{Reporting.}
Each retrained model is trained four times, and we report the mean over the runs. Unless stated otherwise, all numbers are based
on the median statistic.

\noindent\textbf{Reference with labels.}
The ground-truth constant is the global constant fitted with LiDAR ground truth on the validation set using the same rule as $s$
($\times 2.15$ for MonoProb).
It serves as a reference rather than a method, since a label-free result that reaches it achieves what labels provide under a
global constant.

\noindent\textbf{Uncertainty estimates evaluated.}
In addition to MonoProb~\cite{marsal2024_monoprob} and its self-distilled checkpoint, we evaluate the self-supervised estimators
of Poggi~\etal~\cite{poggi2020_uncertainty}, the self-distilled $\sig$ that we trained using the public recipes of
Monodepth2~\cite{godard2019_monodepth2}, MonoViT~\cite{zhao2022_monovit} and Lite-Mono~\cite{zhang2023_litemono}, and three systems
trained with depth labels, namely MaGNet~\cite{bae2022_magnet}, the depth network of OKVIS2-X~\cite{boche2025_okvis2x} and
Metric3D~v2~\cite{hu2024_metric3dv2}. We further use Argoverse~1~\cite{chang2019_argoverse},
DrivingStereo~\cite{yang2019_drivingstereo} and nuScenes~\cite{caesar2020_nuscenes} as additional datasets.

\subsection{Main Results}
\label{sec:exp_main}

\noindent\textbf{Scale.}
As shown in Tab.~\ref{tab:main}, scale calibration reduces the $\rmed$ of MonoProb from 1.94 to 0.96 without labels or retraining,
compared to 0.90 for the ground-truth constant. Since both constants are measured on the validation set, neither yields exactly 1
on the test set.

\noindent\textbf{Shape.}
However, a constant alone is insufficient to correct the shape. It reduces only the part of the consistency error caused by the
scale, while $\sience$ remains unchanged. In contrast, \method{} reduces both the consistency error and $\siencemed$ by a factor of about four
compared to any constant, and both are significantly lower than those of the ground-truth constant
(Tab.~\ref{tab:main}).
On the reliability diagram, the slope increases to 0.95 (Fig.~\ref{fig:teaser}a).
The profile along depth also becomes flatter (Fig.~\ref{fig:profile}). Specifically, the ratio in the nearest range decreases
from about three to about one, whereas a constant alone leaves the uncertainty underestimated in the near range and
overestimated in the middle ranges.
The same conclusions hold for the RMS statistic on consistency and adaptivity.

\noindent\textbf{Standard metrics.}
\method{} also preserves depth accuracy, with an abs\_rel of 0.091 against 0.092 for MonoProb (Tab.~\ref{tab:main}). All seven
standard depth metrics are in Tab.~S3.
AUSE, which is invariant to any constant, is 26\,\% lower than that of MonoProb, although 12\,\% higher than that of the
self-distilled MonoProb checkpoint.
In contrast, ARU and RMSU~\cite{marsal2024_monoprob} increase, which is expected. These metrics are pointwise distances between
$\sig$ and $|e|$, and since $|e|$ is right-skewed, their optima under a global constant are below a calibrated $\sig$.
Seitzer~\etal~\cite{seitzer2022_pitfalls} observed the same effect when using such a distance as a training loss, which leads to an
underestimated variance. Consequently, a $\sig$ that is too small by a factor of two is close to its ARU optimum, and every
calibrated $\sig$ obtains a higher value, by 29--33\,\% for the ground-truth constant and by 8--21\,\% for \method{}
(Tab.~\ref{tab:main}).

\subsection{Generalization}
\label{sec:exp_general}

\begin{figure}[t]
  \centering
  \includegraphics[width=\columnwidth]{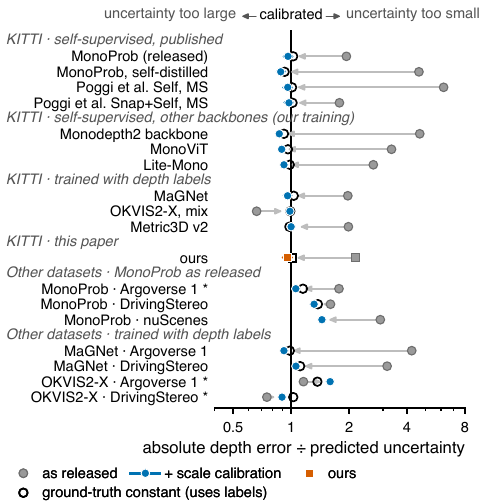}
  \caption{\textbf{Scale calibration matches the ground-truth constant.} Absolute depth error divided by predicted
  uncertainty over all pixels: as released (gray), after scale calibration (blue, with interval or run range) and after the
  ground-truth constant (open). Squares: the retrained $\sig$ of \method{}. $*$: the two constants differ by more than $\times 1.08$.}
  \label{fig:levelall}
\end{figure}

\noindent\textbf{Scale calibration across networks.}
On KITTI, the label-free constant corrects the scale of 17 uncertainty estimates (Fig.~\ref{fig:levelall} shows 11 of them,
Fig.~S1 all). It is within $\times 1.08$ of the ground-truth constant for all of them except the indoor checkpoint of OKVIS2-X.
By construction, the scale with respect to the stereo residual is one, and Fig.~\ref{fig:levelall} shows that it is also one with
respect to the LiDAR error. Scale calibration works in both directions. For example, the uncertainty of the other OKVIS2-X
checkpoint is too large on KITTI (0.66), and scale calibration corrects it to 0.99.

\begin{figure*}[t]
  \centering
  \includegraphics[width=\textwidth]{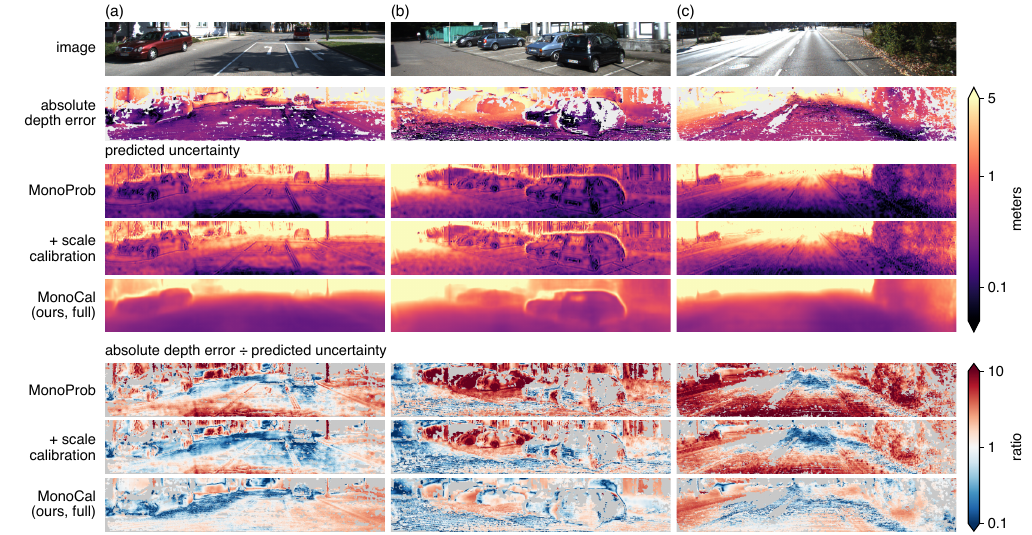}
  \caption{\textbf{Qualitative results.} With \method{}, the ratio of error to uncertainty is closest to 1 (white). Top: on one color
  scale, a calibrated uncertainty is as bright as its error. Bottom: red where the uncertainty is too small, blue where it is too
  large. Three test images.}
  \label{fig:qual}
\end{figure*}

\noindent\textbf{Other datasets.}
On Argoverse~1 and DrivingStereo, where the constant is re-measured, the agreement is weaker than on KITTI, and for
MonoProb neither constant yields a scale of 1 (Fig.~\ref{fig:levelall}).
This is likely due to the stereo depth itself, since semi-global matching produces three times as many errors on DrivingStereo as on
KITTI~\cite{yang2019_drivingstereo}.
Since nuScenes provides no stereo pairs, the constant cannot be measured on this dataset, and reusing the constant measured on
KITTI results in a scale of 1.4 for MonoProb (Fig.~\ref{fig:levelall}).

\begin{table}[t]
  \centering\footnotesize
  \caption{\textbf{Generalization to other backbones and datasets.} Ref.: the backbone's own $\sig$. For consistency it is multiplied by the ground-truth constant that minimizes
  $\encemed$, a stricter reference than the constant of Tab.~\ref{tab:main}. Bottom: the $\sig$ of \method{}, trained on KITTI, with only
  $s$ re-measured, against each dataset's ground-truth constant. Bold: better of each pair.}
  \label{tab:backbones}
  \setlength{\tabcolsep}{7pt}
  \begin{tabular}{l rr rr}
    \toprule
    & \multicolumn{2}{c}{consistency} & \multicolumn{2}{c}{adaptivity} \\
    & \multicolumn{2}{c}{$\encemed$ ($\downarrow$)} & \multicolumn{2}{c}{$\siencemed$ ($\downarrow$)} \\
    \cmidrule(lr){2-3}\cmidrule(lr){4-5}
    backbone / dataset & ref. & ours & ref. & ours \\
    \midrule
    MonoProb   & 0.303 & \textbf{0.077} & 0.194 & \textbf{0.048} \\%
    Monodepth2 & 0.227 & \textbf{0.118} & 0.101 & \textbf{0.059} \\%
    MonoViT    & 0.088 & \textbf{0.072} & \textbf{0.071} & 0.074 \\%
    Lite-Mono  & 0.162 & \textbf{0.077} & 0.098 & \textbf{0.053} \\%
    \midrule
    Argoverse~1   & 0.316 & \textbf{0.167} & 0.278 & \textbf{0.196} \\%
    DrivingStereo & 0.488 & \textbf{0.360} & 0.132 & \textbf{0.053} \\%
    \bottomrule
  \end{tabular}
\end{table}

\noindent\textbf{The disagreement target on other backbones and datasets.}
The disagreement target reduces the shape errors of backbones that exhibit them
(Tab.~\ref{tab:backbones}, Fig.~\ref{fig:profile}). We apply it to Monodepth2, MonoViT and Lite-Mono, each trained using its public
recipe. On Monodepth2 and Lite-Mono, both shape errors decrease. On MonoViT, whose shape errors are already low, only the
consistency error decreases.
On these backbones, AUSE increases by 2--9\,\%.
The second network always has a different architecture from the backbone (Dynamo-Depth~\cite{sun2023_dynamodepth} for MonoViT).
The gain also generalizes to Argoverse~1 and DrivingStereo, where the $\sig$ trained on KITTI, with only $s$ re-measured, achieves
significantly lower shape errors than the ground-truth constant of each dataset.

\subsection{Ablation Study}
\label{sec:exp_ablation}

\begin{table}[t]
  \centering\footnotesize
  \caption{\textbf{Ablation study.} Top: the same training with four sources of the target. Middle: corrections of MonoProb that need no
  training. Bottom: rows that train a $\sig$; \method{} adds the disagreement target to the self-distillation
  of~\cite{marsal2024_monoprob}. Bold: best per block.}
  \label{tab:ablation}
  \setlength{\tabcolsep}{1.3pt}%
  \scriptsize
  \begin{tabular}{l r r r r}
    \toprule
    & $\rmed$ & $\encemed$ & $\ence$ & $\siencemed$ \\
    & ($\to 1$) & ($\downarrow$) & ($\downarrow$) & ($\downarrow$) \\
    \midrule
    \multicolumn{5}{l}{\emph{target of $\sig$, then scale calibration}} \\
    second network (\method{})   & \textbf{0.960} & \textbf{0.077} & \textbf{0.208} & 0.048 \\%
    stereo (SGBM)                & 0.916 & 0.182 & 0.257 & 0.087 \\%
    Depth Anything~V2             & 1.084 & 0.197 & 0.655 & \textbf{0.046} \\%
    \lbl{LiDAR (uses labels)} & \lbl{0.939} & \lbl{0.181} & \lbl{0.311} & \lbl{0.060} \\%
    \midrule
    \multicolumn{5}{l}{\emph{no training: corrections of MonoProb}} \\
    scale calibration                               & 0.964 & 0.301 & 0.969 & 0.194 \\%
    per-$\sig$-bin constants                        & \textbf{0.978} & \textbf{0.056} & 0.912 & 0.138 \\%
    per-($\sig$, depth)-bin constants               & 1.035 & 0.077 & \textbf{0.625} & \textbf{0.089} \\%
    \midrule
    \multicolumn{5}{l}{\emph{one training}} \\
    $\sig$ head only, network frozen                & \textbf{1.032} & 0.098 & 0.377 & 0.064 \\%
    self-distilled~\cite{marsal2024_monoprob} $+$ scale calibration & 0.883 & 0.201 & 0.152 & 0.096 \\%
    self-distilled $+$ per-($\sig$, depth)-bin constants & 0.959 & \textbf{0.051} & \textbf{0.137} & 0.066 \\%
    \addlinespace[1pt]
    \method{} (ours)                                & 0.960 & 0.077 & 0.208 & \textbf{0.048} \\%
    \bottomrule
  \end{tabular}
\end{table}

\noindent\textbf{Choice of target.}
The disagreement target, which uses no labels, achieves a lower consistency error than a LiDAR target.
In the top block of Tab.~\ref{tab:ablation}, we keep the training fixed and vary only the source of the target, using
the second network, stereo, a foundation model~\cite{yang2024_depthanythingv2} and LiDAR.
After scale calibration, the second-network target achieves the lowest consistency error and the second-lowest $\siencemed$.%
These results indicate that the scale is determined by the constant and the shape by the target.

\noindent\textbf{Without training.}
The scale and $\encemed$ can be corrected without training, whereas the adaptivity of \method{} requires it.
We consider three corrections that can be applied to MonoProb without training (Tab.~\ref{tab:ablation}, middle). With scale
calibration alone, both shape errors remain.
Using one constant per $\sig$ bin straightens the reliability diagram, but $\siencemed$ remains high, since the correction
applied per $\sig$ bin is redistributed across depth. This is consistent with the observation of Pernot that consistency does not
imply adaptivity~\cite{pernot2023_adaptivity}.
Using constants per ($\sig$, depth) bin is more effective. These constants are comparable to \method{} on the scale and on
$\encemed$ and reduce $\siencemed$ by half, but $\siencemed$ and $\ence$ remain significantly higher than those of \method{}.

\noindent\textbf{With training.}
Trained alternatives perform close to \method{}, but each at a cost (Tab.~\ref{tab:ablation}, bottom). The self-distillation of
MonoProb is one such training, and \method{} adds the disagreement target to it. With scale calibration alone, the self-distilled
$\sig$ obtains a higher $\encemed$ and a higher $\siencemed$ than \method{} but a lower $\ence$.
Adding constants per ($\sig$, depth) bin makes it comparable to \method{} on KITTI. However, these constants form a table fitted
to KITTI. When reused on Argoverse~1 and DrivingStereo, their $\siencemed$ is 0.242 and 0.094, higher than the 0.196 and 0.053 of
\method{} with only its constant re-measured (Tab.~\ref{tab:backbones}). The full comparison is in Tab.~S8.
The lightest alternative, a $\sig$ head trained on the frozen network with the same target, is worse than \method{} on both
consistency and $\siencemed$.

\subsection{Qualitative Results}
\label{sec:exp_qual}

Fig.~\ref{fig:qual} shows that scale calibration changes how bright the uncertainty is, whereas \method{} changes where it is
bright. In the top rows, the error and the uncertainty share the same color scale in meters, so that a calibrated uncertainty
is as bright as its error, and the uncertainty of MonoProb is visibly darker. The bottom rows show the error divided by the
uncertainty at the LiDAR pixels. For MonoProb, the map is mostly red, indicating that the uncertainty is too small. After scale
calibration, red and blue are balanced, but some regions remain red, such as the upper left of~(b) and the roadsides in~(c), and
\method{} makes these regions closer to white.
The uncertainty of \method{} is also smoother. It no longer follows image texture such as the road markings in~(a), which the error
does not follow either, while preserving the structures where the error is large, such as the vehicles in~(b).

\section{Conclusion and Limitations}
\label{sec:conclusion}

In this paper, we measured the magnitude of the uncertainty of self-supervised monocular depth as the ratio of error to predicted
uncertainty, separating the scale from the shape. We showed that the scale is miscalibrated for every uncertainty estimate we evaluated, by
a factor that varies across networks and datasets. To correct both the scale and the shape without labels, we proposed \method{}.
Scale calibration measures this factor from stereo pairs and requires no retraining, while the disagreement target corrects the
shape, which is unaffected by any constant, by training the uncertainty on the disagreement with a second self-supervised network.

\noindent\textbf{Limitations.}
Although scale calibration is label-free, it still requires stereo pairs to measure the constant. On nuScenes, which provides no
stereo pairs, reusing the constant measured on KITTI results in a scale of 1.4, and on Argoverse~1 and DrivingStereo the constant
is less close to the ground-truth constant than on KITTI.
In addition, the uncertainty is retrained only on KITTI, and on the two other datasets we evaluate the same $\sig$ after
re-measuring the constant.
We leave measuring the constant without a second sensor for future work.

\section*{Acknowledgments}
We thank Laiyan Ding for many discussions that sharpened the research question and the scope of our claims, and for his critical
feedback on the experimental design.

{
    \small
    \bibliographystyle{ieeenat_fullname}
    \bibliography{main}
}

\end{document}